\documentclass[runningheads]{llncs}
\usepackage[T1]{fontenc}
\usepackage{amsmath}
\usepackage{amssymb}
\usepackage{algorithm}
\usepackage{algorithmic}

\usepackage{graphicx}
\usepackage{array}
\usepackage{longtable}
\usepackage{booktabs}
\usepackage{makecell}
\usepackage{tcolorbox}
\usepackage{xcolor}
\usepackage{listings}
\usepackage{color}
\usepackage{hyperref}

\begin{document}
\title{LTLDiff: Finite Linear Temporal Logic–Guided Data Generation and Diffusion Policies for Multi-agent Robotic Manipulation}
\titlerunning{LTLDiff for Multi-Agent Robotic Manipulation}
%
\author{Chuhan Meng\inst{1}\orcidID{0009-0003-9279-6818} \and
Haiyan Yin\inst{2}}

\authorrunning{C. Meng and H. Yin}

\institute{
University of Toronto, Toronto, ON, Canada\\
\email{chuhan.meng@mail.utoronto.ca}
\and
A*STAR Centre for Frontier AI Research (A*STAR CFAR), Agency for Science, Technology and Research (A*STAR), Singapore\\
\email{Yin\_Haiyan@a-star.edu.sg}
}

\maketitle

\begin{abstract}
Multi-agent robotic manipulation tasks require coordination among agents to satisfy task-level temporal, logical, and safety constraints. Recently, diffusion policies have been used to perform the task. However, they still suffer from desynchronization, incorrect action ordering, and coordination failures in tasks that require simultaneous or sequential multi-agent interaction. Therefore, LTLDiff is proposed as a framework that combines Finite Linear Temporal Logic (LTL$_f$) specification learning for both the generation of demonstrations and learning via diffusion policies. Each task has a specific LTL$_f$ formula that is learned from a set of natural language instructions using a large-scale language model. To enable a fixed-dimensional vector embedding of the learned specification from the language model, LTL$_f$ uses an abstract syntax tree representation scheme. This embedding of logic serves as a condition for (i) logic-guided data collection and (ii) diffusion-based policy training, encouraging trajectories that are consistent with the desired ordering and coordination requirements. Experiments on multi-agent LTLDiff manipulation tasks demonstrate improved task success rates compared to the baseline. Together, these contributions demonstrate the effectiveness of LTLDiff for coordinated multi-agent manipulation.

\end{abstract}

\section{Introduction}
Robust developments in robot platforms and control methodologies allow robotic systems to successfully implement tasks with fixed procedures \cite{chi23,zhao23} and carry out tasks following high-level instructions \cite{black24,kim24}. However, applying such capabilities in multi-agent robotics remains an issue. State-of-the-art methods in multi-agent robot control are predominantly found in two categories: one is the rule-based method, and the other is the learning-based method, neither of which has been demonstrated to be sufficient in the following ways. Rule-based methods \cite{boldrer23} enable safety and temporal requirements to be strictly satisfied through predefined logic; However, they suffer from limited flexibility in adapting to novel tasks and changing environmental conditions. The learning-based method \cite{deshpande25} lacks an essential mechanism to satisfy critical safety and temporal requirements in actual task execution, despite having the flexibility to adapt to novel task settings through behavior cloning and reinforcement learning approaches. Recent frameworks such as RoboFactory~\cite{qin25} partially address this challenge by enforcing compositional logical, spatial, and temporal constraints during data generation for multi-agent tasks. Nevertheless, the resulting agents still lack explicit coordination and temporal reasoning during execution because such constraints are not incorporated into policy training.

LTLDiff lies between the rule-based and the learning-based methods. It addresses these complementary limitations by leveraging diffusion policies for task adaptation while incorporating LTL$_f$ specifications as structured conditioning inputs. At the start, LTLDiff utilizes multi-agent robotics coordination by encoding LTL$_f$, which is not enforced as hard rules but rather can be applied generally at runtime with soft gradient guidance. Next, LTLDiff generalizes the logic-guided diffusion policies, LTLDOG~\cite{feng24}, to multi-agent diffusion policies by extending them through the addition of LTL$_f$ conditioning. The gradient guidance is applied during diffusion training by employing a regressor neural network to predict LTL$_f$ satisfaction gradients in the multi-agent scenario, thereby forecasting joint trajectory satisfaction and influencing the bias towards generated LTL$_f$-satisfying behaviors during trajectory sampling. The expressiveness of logic-conditioned algorithms is guided by their dependence on predefined LTL$_f$ specifications. To preserve this structure, LTL$_f$ formulas are parsed into abstract syntax trees and encoded as fixed-dimensional vectors that preserve compositional structure and temporal semantics. As a result, LTLDiff enables reliable conditioning of both the data and diffusion policy on compositional and temporally structured task semantics. Lastly, LTLDiff is applicable to multi-agent execution, where all agents learn simultaneously using different policies and local views. The coordination among the agents occurs implicitly based on the target condition and task instruction, without any joint policy, which shows good scalability. 

The primary contributions of this research work include: (i) An LTL$_f$ generator is proposed to pair LTL$_f$ formulas with each task based on natural-language descriptions of task objectives and guides the data generation process using a large language model. (ii) LTLMAG is proposed for multi-agent regressor-guided planning, which integrates LTL$_f$ constraints into the diffusion policy. (iii)  Experimental results on RoboFactory manipulation tasks demonstrate higher task success rates across most evaluated tasks compared to the baseline method and LTLDOG-R. Hence, they highlight the effectiveness of the proposed LTLDiff framework.

\section{Related Work}
\subsection{Multi-Agent Robotic Systems}
A multi-agent system is composed of several separate agents. Every entity may have conflicting objectives and access to different information. Tool-based agent assistants \cite{wu23autogen} and simulation environments for societies or games \cite{gama2024llmdecision} are the two primary categories of multi-agent systems that are distinguished by their functionalities. In contrast to the aforementioned two, LTLDiff focuses more on the use of multi-agent collaboration, specifically the low-level manipulation of embodied agents.

\subsection{Robot Manipulation}
Traditional robotic manipulation relies on rule-based approaches. A* \cite{foead2021systematic} and LTL-based \cite{banerjee2025review}  are often used to generate collision-free trajectories when the geometric and kinematic constraints are known. These planning and control strategies are restrictive in complex manipulation tasks, despite providing high guarantees of safety and feasibility. Learning-based approaches emphasize the value of flexibility and data-driven management. Behavioral Cloning \cite{dalal2023optimus,jang2022bcz} focuses on demonstrations, and Offline Reinforcement Learning \cite{chebotar2023qtransformer} makes use of offline data. The quality of offline data-based trajectories is further improved by diffusion policies \cite{janner2022diffusion}, which mainly focus on fulfilling safety-related tasks or static constraints without explicitly addressing temporal logic or coordination dependency.

\subsection{Learning and Planning under Temporal Logic}
Finite Linear Temporal Logic is a powerful formalism to express high-level constraints \cite{janner2022diffusion} of tasks in a temporal structure. LTL$_f$ formulas have been applied to robotic planning to specify temporal constraints of tasks \cite{kurtz2023temporal,fainekos2005temporal}. Reinforcement learning in the context of LTL$_f$ objectives was also investigated to efficiently learn policies satisfying LTL$_f$ formulas. The solutions so far require interacting with the physical system through trial-and-error procedures, which may be expensive or even dangerous. Also, the existing solutions using LTL$_f$ are rarely compatible with offline policy learning approaches \cite{vaezipoor2021ltl2action}.
Reliance on trial-and-error interaction is reduced by using LTL$_f$ specifications to guide offline demonstration generation and by incorporating the same logic as a conditioning signal in diffusion-based policy learning.

\section{LTLDiff}
\subsection{Problem Formulation}

LTLDiff focuses on offline learning of multi-agent robotic manipulation with symbolic temporal constraints. The goal is to learn policies that perform physically valid trajectories,  and are guided by LTL$_f$ specifications expressing task-level ordering and coordination requirements.

\begin{figure}[t]
\centering
\IfFileExists{SimpleOverflow.jpg}{%
  \includegraphics[width=0.98\textwidth]{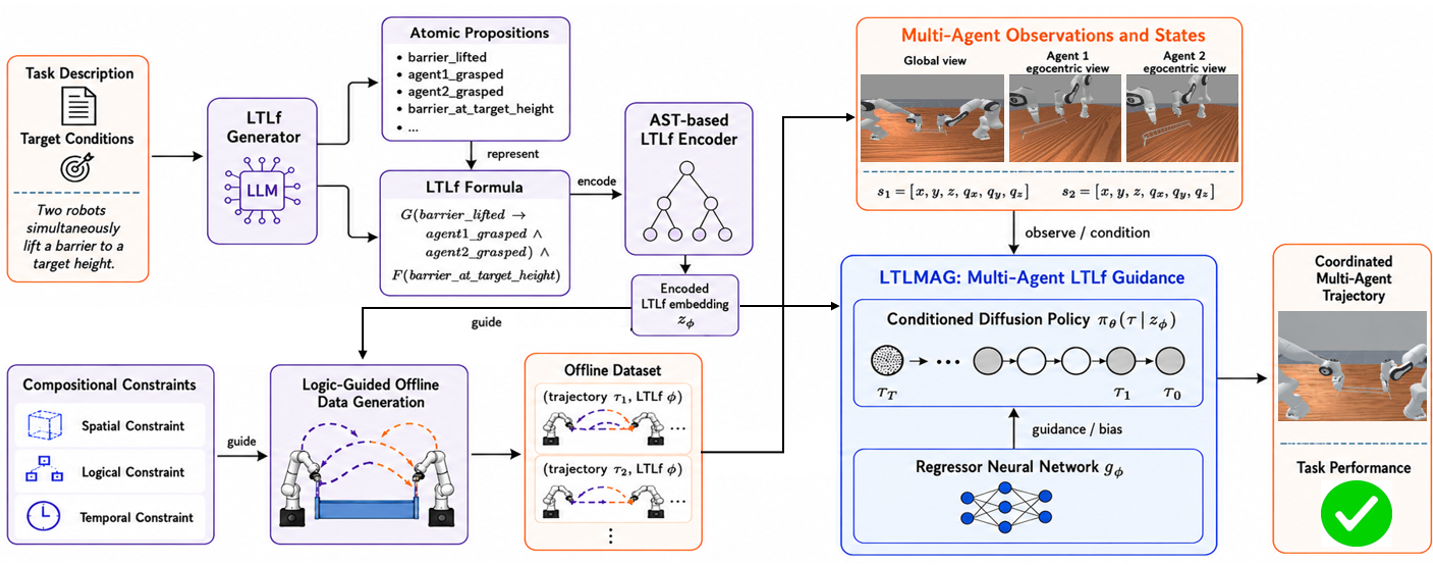}%
}{%
  \fbox{\parbox[c][4.2cm][c]{0.94\textwidth}{\centering\textit{Pipeline figure: SimpleOverflow.jpg}}}%
}
\caption{LTLDiff Pipeline. The task descriptions and conditions are transformed into atomic propositions and an LTL$_f$ formula by an LTL$_f$ generator. The encoded logic and compositional constraints~\cite{qin25} are used in offline data generation, whereas multi-agent observations and the LTL$_f$ embedding are utilized by LTLMAG for coordinated trajectory generation.}
\label{fig1}
\end{figure}

\subsubsection{Multi-Agent Setting.}
A system of $N$ robotic agents $\{a_1, \ldots, a_N\}$ is operating in a shared
environment. Agents receive RGB observations
$\mathcal{O} = \{o_{\text{global}}, o_1, \ldots, o_N\}$, where $o_{\text{global}}$ denotes
a global scene view and $o_i$ denotes the egocentric RGB observation of agent $i$. At each time step $t$, agent $a_i$ receives an observation
$o_i^t \in \mathcal{O}$, consisting of visual observations and states, $s_i^t \in \mathcal{S}$. A joint multi-agent trajectory $\mathcal{\tau}$ over a
finite horizon $T$ is defined as:

\begin{equation}
\tau = \left\{\left(o_{\text{global}}^t, o_1^t, \ldots, o_N^t,
 s_1^t, \ldots, s_N^t\right)\right\}_{t}^{T}.
\end{equation}

\subsubsection{LTL$_f$ Task Specification.}
Each task is associated with a finite-horizon LTL formula $\varphi$ defined over a set of
atomic propositions $\mathcal{P}$. These propositions correspond to task-relevant events
such as objects, names, agents, etc. The LTL$_f$ formula is formed according to the task description and target condition. There are some temporal and logical constraints included implicitly in the description, such as ordering (e.g., ``grasp before lift''), safety (e.g., ``always avoid collision''), or coordination (e.g., ``both agents must grasp
before lifting'').

\subsubsection{Offline Dataset.}
An offline dataset is formed by using:
\begin{equation}
\mathcal{D} = \{(\tau^{(k)}, \varphi)\}_{k=1}^{K},
\label{dataset}
\end{equation}
where each demonstration trajectory $\tau^{(k)}$ is paired with its task LTL$_f$
specification $\varphi$. The dataset is collected before training, and no
environment interaction during learning.

\subsubsection{Learning Objective.}
The aim of LTLDiff is to train a conditional diffusion policy to execute tasks.
\begin{equation}
\pi_{\theta}(\tau \mid \mathbf{z}_{\varphi}),
\end{equation}
$\pi_{\theta}$ represents the diffusion policy and
$\mathbf{z}_\varphi \in \mathbb{R}^d$ represents the LTL$_f$ formula after encoding it into
a fixed-dimensional vector, which is passed to the diffusion policy as a conditioning signal.

\subsection{LTL$_f$ Generator}
Each manipulation task begins with a natural language description of its task and target conditions. These descriptions implicitly state temporal ordering and logical dependencies that must be extracted and formalized. A large language model called Qwen is used to generate the LTL$_f$. Each LTL$_f$ formula is generated ten times using Qwen. Qwen then re-examines the task description and target condition to select the most appropriate formula from the ten candidates. The task description, target condition, and full prompt are provided in the supplementary material.

These generated formulas are presented in a formatted JSON structure, which includes a description of the tasks, a set of atomic statements, and an LTL$_f$ formula defined in the statements. Every atomic statement can be viewed as an expression in Boolean manipulation based on an event or a state in a manipulation task. It represents different states associated with objects, interactions between agents and objects, or coordination events.

The LTL$_f$ is built through a restricted set of temporal and logical operators consistent with the implementation. Temporal relationships are expressed using the operators \texttt{F} (eventually), \texttt{G} (globally), \texttt{U} (until), and
\texttt{X} (next), which capture ordering constraints, safety invariants, and goal conditions
over time. Logical structure is represented using the operators \texttt{\&} (and),
\texttt{|} (or), \texttt{->} (implication), and \texttt{!} (negation), allowing the specification of conditional dependencies and forbidden conditions.

\subsection{LTL$_f$ Encoder}

After an LTL$_f$ specification is generated, it is converted to a structured 128-dimensional vector representation by using the LTL$_f$ Encoder. This format can be passed to the data generation process and diffusion policy. The encoding process has two stages: tokenization and parsing, and AST-based embedding.

\subsubsection{Tokenization and Parsing.}
Operators, propositions, and parentheses are explicitly separated to tokenize the LTL$_f$ formula  $\varphi \in \mathrm{LTL}_f(P)$, where $P$ is the set of atomic propositions defined for the task. The tokenized formula is then parsed to generate an Abstract Syntax Tree (AST) to denote its hierarchical structure. The logical combination expression and the temporal constraints conveyed in the LTL$_f$ formula are preserved in the AST.

\subsubsection{AST-based Encoding.}
To enable learning with fixed-size neural network inputs, the Abstract Syntax Tree (AST) $\mathcal{T}_\varphi = (V, E)$ is converted into a fixed-dimensional vector representation via recursive encoding. Each node $v \in V$ corresponds to either an atomic proposition (leaf node) $p \in P$ or a temporal or logical operator (internal node) $o \in \{F, G, X, U, \texttt{\&}, \texttt{|}, \texttt{-\!>}, \texttt{!}\}$. The edge set $E$ defines the parent--child relationships between operators and their operands, preserving the hierarchical structure of the LTL$_f$ formula. The resulting root embedding provides a compact representation of the full specification.

Every leaf node in the AST is encoded as a one-hot vector on the set of propositions and represents an atomic proposition. For a leaf node $v$ corresponding to an atomic proposition $p$, the node embedding is defined as
\begin{equation}
\mathbf{h}_v = \text{OneHot}(p).
\end{equation} 
The encoding process for an internal node starts with its operator encoding, which combines with the encoded child nodes $\mathbf{h}_u$ to show the internal structure of the node. The operator encoding uses a summary representation which results from recursive computation and averaging of embedded child nodes $\mathcal{C}(v)$ to create their combined encoding. Equation 5 defines the internal node:
\begin{equation}
\mathbf{h}_v =
\Big[
\text{OneHot}(o_v)
\;\Vert\;
\frac{1}{|\mathcal{C}(v)|}
\sum_{u \in \mathcal{C}(v)} \mathbf{h}_u
\Big].
\end{equation}
where $\text{OneHot}(o_v)$ denotes the one-hot encoding of the operator type at node $v$.
Finally, the embedding of the full LTL$_f$ formula is given by the root node of the AST:
\begin{equation}
\mathbf{z}_\varphi =
\text{Pad}\!\left(\mathbf{h}_{\text{root}(\mathcal{T}_\varphi)}\right)
\in \mathbb{R}^d,
\end{equation}
where $\text{Pad}(\cdot)$ denotes zero-padding to a fixed dimension $d = 128$, consistent with the regressor and diffusion policy input size.

\subsection{LTL$_f$ Guided Offline Data Generation}

With the aim of producing demonstrations that are applicable to the task, task-level logic is incorporated into the motion planner employed in the task demonstration data generation process. In this regard, each task is considered to be coupled with a fixed solution structure, which is constructed by a set of motion primitives. In addition, the data generation process is divided into various phases, which are determined by the atomic propositions and the temporal structure defined in the LTL$_f$ formula. A phase is defined as every semantically valid sub-task, according to the logic constraints. The atomic propositions are used as the interface between high-level task semantics and low-level motion execution, specifying the active and relevant task aspects. 

A structured vector representation of the LTL$_f$ formula is formed by the atomic propositions, and it is used as a conditioning bias for modulating the motion planning parameters. These motion planning parameters include velocity, smoothness of accelerations, and object position. Rather than directly mapping atomic propositions to specific parameter values, the presence of particular propositions in the LTL$_f$ formula determines which motion planning parameters are eligible for modulation, while the corresponding LTL$_f$ embedding values determine the magnitude and direction of the applied adjustment. The motion planner utilizes this conditioning bias for generating motion trajectories, which are more correlated with the temporal structure of the LTL$_f$ formula. It generates demonstrations in the form of pairs made up of a trajectory and its LTL$_f$ formula specification, defined in Eq.~\eqref{dataset}.

Although RoboFactory supports spatial, temporal, and logical constraints through separate modules that operate during data generation, these constraints do not have a unified mechanism to specify task-level semantics. However, LTL$_f$ has a unified compositional form with explicit temporal and logical operators, which is beneficial for reasoning over entire trajectories and ensuring that local behavior is not globally inconsistent. Therefore, LTL$_f$ is a machine-interpretable formalism that unifies multi-agent interaction modeling, formal verifiability, and conditioning for logic-guided data generation.

\subsection{LTLMAG: Multi-Agent LTL$_f$ Guidance}
\label{3.5}
\begin{algorithm}
  \caption{LTLMAG: Multi-Agent LTL$_f$ Planning with Regressor Guidance}
  \label{alg:single_ltlf_planning}
  \begin{algorithmic}
    \STATE {\bfseries Require:} diffusion steps $N$, agent number $N_a$, guidance signal $\xi_i$, regressor neural network $g_\phi(\cdot)$
    \STATE Initialize $\tau^{N} \sim \mathcal{N}(0, I)$
    \FOR{$i = N-1$ {\bfseries down to} $1$}
        \FOR{$n = 1$ {\bfseries to} $N_a$}
        \STATE Sample noise $\epsilon \sim \mathcal{N}(0, I)$
        \STATE $\tau_n^{i-1} \leftarrow 
        \frac{1}{\sqrt{\alpha_i}}
        \Big(\tau_n^{i} + (1-\alpha_i)\,
        \pi_\theta(\tau_n^{i} \mid \mathbf{z}_{\varphi})\Big)
        + \sqrt{1-\alpha_i}\,\epsilon$

        \ENDFOR
        \STATE $\tau_{\text{all}}^{i-1} \leftarrow \operatorname{concat}\!\left(\{\tau_n^{i-1}\}_{n=1}^{N_a}\right)$

        \STATE $\tau_{\text{all}}^{i-1} \leftarrow \tau_{\text{all}}^{i-1}
        + \xi_i\, g_\phi(\tau_{\text{all}}^{i}, \mathbf{z}_{\varphi}, i)$

    \ENDFOR
    \STATE {\bfseries return} $\tau^{0}_{\text{all}}$
  \end{algorithmic}
\label{LTLMAG}
\end{algorithm}

LTLMAG is used as the multi-agent LTL$_f$ Guidance System, which is shown in Algorithm~\ref{LTLMAG}. At the beginning, the diffusion policy starts with Gaussian noise and gradually improves trajectories in the direction of workable solutions. Next, to create action trajectories for each agent, LTLMAG uses the conditioned diffusion policy in which sampling is done via an iterative reverse denoising procedure. A single LTL$_f$ specification defines the temporal and coordination constraints of the task and is shared across all agents. This specification is provided as input to both the conditioned diffusion policy $\pi_\theta$ and a learned regressor network $g_\phi$. The regressor is trained to approximate the target LTL$_f$ satisfaction gradient. Let $g_{\phi}(\tau_t,z_{\varphi},t)$ denote the correction predicted by the regressor, and let $g^{*}(\tau_t,z_{\varphi})$ denote the target gradient direction obtained from the satisfaction estimator. The training loss is defined as

\begin{equation}
\mathcal{L}_{\mathrm{reg}}
=
\left\|
g_{\phi}
-
g^{*}
\right\|_2^2
+
0.01
\max
\left(
0,
\left\|
g_{\phi}
\right\|_2
-
1
\right)^2
+
0.1\mathcal{L}_{\mathrm{sat}} .
\end{equation}
where $\mathcal{L}_{\mathrm{sat}}$ is an auxiliary binary cross-entropy loss for predicting LTL$_f$ satisfaction when satisfaction labels are available. During the reverse diffusion process, the output of the regressor neural network and the constant guidance signal $\xi_i$ bias the trajectory generation toward satisfying the symbolic task requirements in the denoising dynamics. Here, $\xi_i$ is a time-varying guidance weight that controls the strength of LTL$_f$ satisfaction guidance at timestep $i$. By default, the value is set to 0.1. The guidance follows a linear schedule, with stronger guidance at early timesteps $(2\xi_i)$ and weaker guidance at later timesteps $(0.1\xi_i)$. The guidance weight is fixed across agents and tasks, but varies across diffusion timesteps. Larger values place stronger emphasis on satisfying the LTL$_f$ formula, while smaller values make the sampler rely more on the learned diffusion prior. Each agent keeps its path and performs local updates using reverse diffusion independently. The per-agent trajectories are concatenated at each cycle to form a joint multi-agent trajectory, which represents the clean path predictions of all agents after local denoising.

Three key improvements are introduced to LTLDOG-R \cite{feng24} that enable the derivation of LTLMAG. First, a conditioned diffusion policy is applied to generate task-aware trajectories, considering both observations and LTL$_f$ embeddings through global conditioning. The reason is that unconditional diffusion policy generates task-agnostic trajectories and depend on external guidance to enforce coordination, which is unreliable at early denoising stages due to high noise levels. Conditioning the Diffusion Policy on LTL$_f$ embeddings reshapes the generative prior, steering denoising toward task-consistent and coordinated trajectories from the outset. This conditioning allows the diffusion policy to internalize temporal structure during training, reducing the need for strong guidance at inference and improving stability and sample efficiency in long-horizon multi-agent tasks. Second, the original Graph Convolutional Neural Network is replaced with a feed-forward neural network for the LTL$_f$ regressor, which is the guidance signal. Since robot trajectories take the form of flattened action sequences, multilayer perceptrons appear to be an efficient and effective method for approximating functions that compute satisfaction scores for trajectory–constraint pairs and their gradient. Third, the algorithm is extended from single-agent to multi-agent through a two-stage update: per-agent local denoising within each diffusion step, followed by regressor-based guidance applied to the joint multi-agent trajectory to enforce cross-agent temporal and coordination constraints. This extension enables coordinated trajectory generation and joint logic satisfaction across multiple interacting agents.
This is because LTLDOG-R was originally defined for single-agent problems only, LTLMAG algorithm will support multi-agent problems as demanded within RoboFactory.

\section{Experiment}

\begin{table}
\caption{Task success rates (\%) across the baseline, LTLDOG-R (LTLR), and LTLDiff under different dataset sizes.}
\label{tab:success-rate}
\centering
\resizebox{\textwidth}{!}{
\begin{tabular}{llccccccccc}
\hline
\multicolumn{2}{c}{} 
& \multicolumn{3}{c}{50 Demos} 
& \multicolumn{3}{c}{100 Demos} 
& \multicolumn{3}{c}{150 Demos} \\
\hline
Task Level & Task Name
& Base & LTLR & LTLDiff
& Base & LTLR & LTLDiff
& Base & LTLR & LTLDiff \\
\hline

1-Agent
& Pick Meat   & 32 & 65 & 80 & 61 & 82 & 93 & 58 & 76 & 87 \\
& Stack Cube  & 17 & 13 & 9  & 38 & 27 & 18 & 44 & 32 & 21 \\
& Strike Cube & 26 & 66 & 80 & 42 & 55 & 64 & 40 & 62 & 75 \\
& Average     & 25 & 48 & \underline{\textbf{56}}
              & 47 & 55 & \underline{\textbf{58}}
              & 47 & 57 & \underline{\textbf{61}} \\
\hline

2-Agent
& Pass Shoe             & 9  & 5  & 0  & 20 & 12 & 3  & 12 & 9  & 7  \\
& Place Food            & 5  & 18 & 25 & 23 & 23 & 23 & 20 & 23 & 25 \\
& Lift Barrier          & 24 & 67 & 81 & 60 & 82 & 92 & 58 & 84 & 96 \\
& Two Robots Stack Cube & 14 & 22 & 27 & 27 & 32 & 35 & 20 & 48 & 60 \\
& Average               & 13 & 28 & \underline{\textbf{33}}
                        & 33 & 37 & \underline{\textbf{38}}
                        & 28 & 41 & \underline{\textbf{47}} \\
\hline

3-Agent
& Camera Alignment        & 7 & 45 & 56 & 10 & 60 & 73 & 19 & 55 & 67 \\
& Three Robots Stack Cube & 8 & 9  & 10 & 2  & 3  & 3  & 22 & 14 & 7  \\
& Average                 & 8 & 27 & \underline{\textbf{33}}
                          & 6 & 32 & \underline{\textbf{38}}
                          & 21 & 35 & \underline{\textbf{37}} \\
\hline

4-Agent
& Take Photo             & 5 & 9  & 12 & 8 & 10 & 12 & 20 & 16 & 24 \\
& Long Pipeline Delivery & 0 & 0  & 0  & 0 & 0  & 0  & 0  & 0  & 0  \\
& Average                & 3 & 5 & \underline{\textbf{6}}
                        & 4 & 5 & \underline{\textbf{6}}
                        & 10 & 8 & \underline{\textbf{12}} \\
\hline
\end{tabular}
}
\end{table}

The experiments consist of two parts: (i) a comparison of task performance among LTLDiff, the RoboFactory baseline, and the LTLDOG-R; (ii) an ablation study on the use of LTL$_f$ formulas in data generation and diffusion policy training.

\subsection{Comparative Evaluation of LTLDiff and the RoboFactory Baseline}
\label{4.1}

Conditioned diffusion policies are trained with egocentric RGB observations in an offline imitation learning setting. A total of 11 tasks are evaluated; for each task, policies are trained independently using 50, 100, and 150 expert demonstrations. The baseline diffusion policy is just a diffusion model. LTLDOG-R uses regressor guidance and is integrated into multi-agent problems. LTLDiff is improved over LTLDOG-R, and the improvements are described in Section~\ref{3.5}. Three policies are trained under identical training settings, and the comparative results are reported in Table~\ref{tab:success-rate}.

\subsubsection{Single-agent Tasks.}
The average task success rate of demos all improved significantly over the baseline and the LTLDOG-R. Although the result shows that the Stack Cube task is particularly challenging, it highlights the difficulty of accurately manipulating objects based solely on visual observations, as well as the impact of the height of the cube on its bouncing or rolling behavior. However, LTLDiff performs better as the number of learning demonstrations increases. This improvement is attributed to LTL$_f$ conditioning, which provides appropriate inductive bias even in tasks that require less coordination.

\subsubsection{Two-agent Tasks.}
Increases in terms of performance are more evident in the two-agent tasks. There are marked improvements in the performance on the Lift Barrier and Two Robots Stack Cube problems in the 150-demo case. On average, across all two-agent tasks, LTLDiff achieves the highest success rate.

\subsubsection{Three-agent Tasks.}
LTLDiff shows a large improvement in the three-agent tasks. The average success rate of the 50, 100, and 150 demos increased by 25\%, 32\%, and 16\%, respectively, compared to the baseline.

\subsubsection{Four-agent Tasks.}
Although the average success rate for all demos is higher than the baseline and the LTLDOG-R, it still remains low due to the complexity of coordinating over a long-term horizon and adhering to overall temporal constraints. Specifically, the "Long Pipeline Delivery" problem needs the learning system to maintain consistent temporal dependencies over long-term horizon constraints, which has been quite challenging for the diffusion-based policy learning method from demonstrations. 

In conclusion, the results suggest that LTL$_f$ conditioning plays a more significant role in improving task success than the regressor neural network alone.

\subsection{Ablation Study}
\begin{table}
\caption{Task success rates (\%) for ablation studies of LTL$_f$. The experiment (Gen) is performed under the same conditions in Section~\ref{4.1}, but only with LTL$_f$ in the data generation. The experiment (Train) is performed in which all LTL$_f$ constraints remain in diffusion only, and all other setups are the same.}
\label{tab:ablation}
\centering
\begin{tabular}{llcccccc}
\hline
\multicolumn{2}{c}{} 
& \multicolumn{2}{c}{50 Demos} 
& \multicolumn{2}{c}{100 Demos} 
& \multicolumn{2}{c}{150 Demos} \\
\hline
Task Level & Task Name
& Gen & Train
& Gen & Train
& Gen & Train \\
\hline

1-Agent
& Pick Meat   & 45 & 74 & 72 & 86 & 68 & 80 \\
& Stack Cube  & 6 & 7  & 10 & 15 & 12 & 20 \\
& Strike Cube & 55 & 74 & 52 & 59 & 60 & 69 \\
& Average     & 35 & \underline{\textbf{52}}
              & 45 & \underline{\textbf{53}}
              & 47 & \underline{\textbf{56}} \\
\hline

2-Agent
& Pass Shoe             & 1  & 0  & 5  & 3  & 10  & 7  \\
& Place Food            & 15 & 23 & 20 & 22 & 23 & 23 \\
& Lift Barrier          & 60 & 75 & 70 & 85 & 75 & 89 \\
& Two Robots Stack Cube & 18 & 25 & 30 & 33 & 45 & 56 \\
& Average               & 24 & \underline{\textbf{31}}
                        & 31   & \underline{\textbf{36}}
                        & 38   & \underline{\textbf{44}} \\
\hline

3-Agent
& Camera Alignment        & 10 & 52 & 40 & 67 & 55 & 62 \\
& Three Robots Stack Cube & 8  & 9  & 1  & 3  & 2  & 7  \\
& Average                 & 9 & \underline{\textbf{31}}
                          & 21 & \underline{\textbf{35}}
                          & 29 & \underline{\textbf{35}} \\
\hline

4-Agent
& Take Photo             & 6 & 9 & 6 & 11 & 7 & 12 \\
& Long Pipeline Delivery & 0 & 0  & 0 & 0  & 0 & 0  \\
& Average                & 3 & \underline{\textbf{5}}
                        & 3 & \underline{\textbf{6}}
                        & 4 & \underline{\textbf{6}} \\
\hline
\end{tabular}
\end{table}
The ablation study examines the importance of LTL$_f$ within LTLDiff by addressing the following questions: (1) What is the importance of LTL$_f$ constraints in the data generation process? (2) What is the importance of LTL$_f$ constraints in diffusion policy training? To answer these questions, two experiments are conducted, and the results are shown in Table~\ref{tab:ablation}.

\subsubsection{Importance of LTL$_f$ in Data Generation}
Incorporating LTL$_f$ constraints into data generation captures task-level temporal logic that is difficult to infer from demonstrations alone. Without constraints from the symbolic environment, the diffusion policy inherits correlations between actions based solely on feasibility, resulting in repeated coordination errors in multi-agent settings. For example, agents may act asynchronously in parallel tasks (e.g., Camera Alignment) or execute incorrect action orders (e.g., Stacking tasks), leading to coordination failures.

\subsubsection{Importance of LTL$_f$ in Diffusion Policy}

The diffusion policies are conditioned on LTL$_f$ to bias trajectory sampling based on behaviors that satisfy the desired temporal relationships. Although diffusion models can learn the correlation between local actions, their behavior can become unpredictable with the increase in task complexity due to the lack of explicit global structure in the task. This can cause collisions, unnecessary actions, and inefficient synchronization of the agents, especially in three- and four-agent tasks, as ablation results show a significant decrease in success rate compared to LTLDiff. The diffusion process is given a crucial inductive bias through the conditioning of LTL$_f$ that favors temporal consistency and coordinated behavior, especially in complex multi-agent tasks.

\section{Conclusion}
This work investigates the usage of LTL$_f$ in the multi-agent robotic manipulation with regard to learning from demonstrations. Task level constraints are integrated during the generation of demonstrations. In particular, demonstrations are generated with the aid of motion planners and LTL$_f$ formulas. The demonstrations are then used to train diffusion policies, which can learn task-level temporal properties without the need for hard constraint enforcement during execution. Based on the experiments conducted on the RoboFactory benchmark, diffusion policies learned with constraint-aware demonstrations outperformed diffusion policies on most of the multi-agent manipulation tasks with a high level of coordination.

Despite these contributions, LTLDiff still has a limitation: its performance is poor when processing long-horizon tasks. To solve this problem, in the future, incorporate a closed-loop feedback, such as receding horizon control, which enables the system to continually replan based on real-time observations. This solution enhances robustness by correcting execution errors and compensating for state distributions not represented in the demonstration dataset.

%
%

\begin{thebibliography}{8}


\bibitem{chi23}
Chi, C., Xu, Z., Xu, S., Yu, T., Wang, J., Li, Y.:
Diffusion policy: Visuomotor policy learning via action diffusion.
arXiv preprint arXiv:2303.04137 (2023)

\bibitem{zhao23}
Zhao, T.Z., Kumar, V., Levine, S., Finn, C.:
Learning fine-grained bimanual manipulation with low-cost hardware.
arXiv preprint arXiv:2304.13705 (2023)

\bibitem{black24}
Black, K., Brown, N., Driess, D., Liu, C., Lynch, C., Ryoo, M.S., Levine, S.:
$\pi_0$: A vision-language-action flow model for general robot control.
arXiv preprint arXiv:2410.24164 (2024)

\bibitem{kim24}
Kim, M.J., Pertsch, K., Karamcheti, S., Xiao, T., Balakrishna, A.,
Nair, S., Rafailov, R., Foster, E., Lam, G., Sanketi, P.,
Vuong, Q., Kollar, T., Burchfiel, B., Tedrake, R., Sadigh, D.,
Levine, S., Finn, C.:
OpenVLA: An Open-Source Vision-Language-Action Model.
arXiv preprint arXiv:2406.09246 (2024)

\bibitem{feng24}
Feng, Z., Luan, H., Goyal, P., Soh, H.:
LTLDOG: Satisfying temporally-extended symbolic constraints for safe diffusion-based planning.
IEEE Robotics and Automation Letters \textbf{9}(10), 8571--8578 (2024)

\bibitem{qin25}
Qin, Y., Kang, L., Song, X., Yin, Z., Liu, X., Liu, X.,
Zhang, R., Bai, L.:
RoboFactory: Exploring Embodied Agent Collaboration with Compositional Constraints.
arXiv preprint arXiv:2503.16408 (2025)

\bibitem{deshpande25}
Deshpande, N., Vaidya, A.S., Samant, R.C., Mishra, P., Biradar, V., Keerthi, P.:
Multi-agent deep reinforcement learning for multi-robot systems:
A survey of challenges and applications.
International Journal of Environmental Sciences \textbf{11}(4s), 111--116 (2025)

\bibitem{boldrer23}
Boldrer, M., Serra-Gomez, A., Lyons, L., Kratky, V., Alonso-Mora, J., Ferranti, L.:
Rule-based Lloyd algorithm for multi-robot motion planning and control with safety and convergence guarantees.
arXiv preprint arXiv:2310.19511 (2023)

\bibitem{foead2021systematic}
Foead, D., Ghifari, A., Kusuma, M.B., Hanafiah, N., Gunawan, E.:
A systematic literature review of A* pathfinding.
Procedia Computer Science \textbf{179}, 507--514 (2021)

\bibitem{banerjee2025review}
Banerjee, A.S., Choppella, V.:
Challenges and Opportunities in the Industrial Usage Controller Synthesis Tools:
A Review of LTL-Based Opensource Tools for Automated Control Design.
Results in Control and Optimization 18, 100511 (2025).

\bibitem{wu23autogen}
Wu, Q., Bansal, G., Zhang, J., Wu, Y., Li, B., Zhu, E., Jiang, L., Zhang, X., Zhang, S., Liu, J., Awadallah, A.H., White, R.W., Burger, D., Wang, C.:
AutoGen: Enabling next-gen LLM applications via multi-agent conversation.
arXiv preprint arXiv:2308.08155 (2023)

\bibitem{gama2024llmdecision}
Huang, J.-t., Li, E.J., Lam, M.H., Liang, T., Wang, W., Yuan, Y.,
Jiao, W., Wang, X., Tu, Z., Lyu, M.R.:
How Far Are We on the Decision-Making of LLMs? Evaluating LLMs'
Gaming Ability in Multi-Agent Environments.
arXiv preprint arXiv:2403.11807 (2024)

\bibitem{dalal2023optimus}
Dalal, M., Mandlekar, A., Garrett, C., Handa, A., Salakhutdinov, R., Fox, D.:
Imitating task and motion planning with visuomotor transformers.
arXiv preprint arXiv:2305.16309 (2023)

\bibitem{janner2022diffusion}
Janner, M., Du, Y., Tenenbaum, J.B., Levine, S.:
Planning with diffusion for flexible behavior synthesis.
In: Proceedings of the 39th International Conference on Machine Learning (ICML 2022),
pp. 9902--9915. PMLR (2022)


\bibitem{jang2022bcz}
Jang, E., Irpan, A., Khansari, M., Kappler, D., Ebert, F., Lynch, C., Levine, S., Finn, C.:
BC-Z: Zero-shot task generalization with robotic imitation learning.
In: Proc. Conf. on Robot Learning (CoRL), pp. 991--1002. PMLR (2022)

\bibitem{chebotar2023qtransformer}
Chebotar, Y., Vuong, Q., Hausman, K., Xia, F., Lu, Y., Irpan, A., Kumar, A., Yu, T., Herzog, A., Pertsch, K., et al.:
Q-Transformer: Scalable offline reinforcement learning via autoregressive Q-functions.
In: Proc. Conf. on Robot Learning (CoRL), pp. 3909--3928. PMLR (2023)

\bibitem{kurtz2023temporal}
Kurtz, V., Lin, H.:
Temporal logic motion planning with convex optimization via graphs of convex sets.
IEEE Transactions on Robotics \textbf{39}(5), 3791--3804 (2023)



\bibitem{fainekos2005temporal}
Fainekos, G.E., Kress-Gazit, H., Pappas, G.J.:
Temporal logic motion planning for mobile robots.
In: Proceedings of the IEEE International Conference on Robotics and Automation (ICRA 2005),
pp. 2020--2025. IEEE (2005)

\bibitem{vaezipoor2021ltl2action}
Vaezipoor, P., Li, A.C., Icarte, R.A.T., McIlraith, S.A.:
LTL2Action: Generalizing LTL instructions for multi-task reinforcement learning.
In: Proceedings of the 38th International Conference on Machine Learning (ICML 2021),
Proceedings of Machine Learning Research, pp. 10497--10508 (2021)

\end{thebibliography}
%

\appendix
\sloppy
\section{LTL$_f$ Formulae}

\begingroup
\scriptsize
\setlength{\tabcolsep}{3pt}
\renewcommand{\arraystretch}{1.15}

\begin{longtable}{@{}p{0.16\textwidth} p{0.37\textwidth} p{0.47\textwidth}@{}}
\caption{LTL$_f$ formulas for each robotic manipulation task.}
\label{tab:ltl_formulas}\\

\toprule
\textbf{Task} & \textbf{Atomic Propositions} & \textbf{LTL Formula} \\
\midrule
\endfirsthead

\toprule
\textbf{Task} & \textbf{Atomic Propositions} & \textbf{LTL Formula} \\
\midrule
\endhead

\bottomrule
\endfoot

Pick Meat &
\makecell[l]{
$p_1$: meat grasped\\
$p_2$: meat lifted\\
$p_3$: meat reaches target height
}
&
\makecell[l]{
$\mathbf{G}(\mathord{!}p_3 -> (\mathord{!}p_3\ \mathbf{U}\ p_1))$\\
$\&\ \mathbf{G}(p_1 -> \mathbf{X}(\mathbf{F}p_2))$\\
$\&\ \mathbf{G}(p_2 -> \mathbf{F}p_3)$
}
\\

\midrule

Stack Cube &
\makecell[l]{
$p_1$: blue cube grasped\\
$p_2$: blue cube placed on red cube\\
$p_3$: blue and red cubes aligned\\
$p_4$: blue cube above red cube
}
&
\makecell[l]{
$\mathbf{G}(\mathord{!}p_2 -> (\mathord{!}p_2\ \mathbf{U}\ p_1))$\\
$\&\ \mathbf{G}(p_1 -> \mathbf{X}(\mathbf{F}p_2))$\\
$\&\ \mathbf{G}(p_2 -> (p_3\ \&\ p_4))$
}
\\

\midrule

Strike Cube &
\makecell[l]{
$p_1$: hammer identified\\
$p_2$: hammer grasped\\
$p_3$: hammer above cube\\
$p_4$: cube struck
}
&
\makecell[l]{
$\mathbf{G}(\mathord{!}p_4 -> (\mathord{!}p_4\ \mathbf{U}\ p_1))$\\
$\&\ \mathbf{G}(p_1 -> \mathbf{X}(\mathbf{F}p_2))$\\
$\&\ \mathbf{G}(p_2 -> \mathbf{F}(p_3\ \&\ \mathbf{F}p_4))$\\
$\&\ \mathbf{G}(p_4 -> p_3)$
}
\\

\midrule

Lift Barrier &
\makecell[l]{
$p_1$: agent 1 grasps barrier\\
$p_2$: agent 2 grasps barrier\\
$p_3$: both agents grasp barrier\\
$p_4$: barrier reaches target height\\
$p_5$: barrier remains stable
}
&
\makecell[l]{
$\mathbf{G}((p_1\ \&\ p_2) -> p_3)$\\
$\&\ \mathbf{G}(\mathord{!}p_4 -> (\mathord{!}p_4\ \mathbf{U}\ p_3))$\\
$\&\ \mathbf{G}(p_3 -> \mathbf{X}(\mathbf{F}(p_4\ \&\ p_5)))$\\
$\&\ \mathbf{G}(p_4 -> p_5)$
}
\\

\midrule

Pass Shoe &
\makecell[l]{
$p_1$: first arm grasps shoe\\
$p_2$: shoe transferred to second arm\\
$p_3$: second arm grasps shoe\\
$p_4$: shoe reaches target location
}
&
\makecell[l]{
$\mathbf{G}(\mathord{!}p_4 -> (\mathord{!}p_4\ \mathbf{U}\ p_1))$\\
$\&\ \mathbf{G}(p_1 -> \mathbf{X}(\mathbf{F}(p_2\ \&\ p_3)))$\\
$\&\ \mathbf{G}((p_2\ \&\ p_3) -> \mathbf{F}p_4)$\\
$\&\ \mathbf{G}(p_4 -> p_3)$
}
\\

\midrule

Place Food &
\makecell[l]{
$p_1$: pot lid opened\\
$p_2$: food grasped\\
$p_3$: food placed inside pot\\
$p_4$: food near pot center
}
&
\makecell[l]{
$\mathbf{G}(\mathord{!}p_3 -> (\mathord{!}p_3\ \mathbf{U}\ (p_1\ |\ p_2)))$\\
$\&\ \mathbf{G}((p_1\ |\ p_2) -> \mathbf{F}(p_1\ \&\ p_2))$\\
$\&\ \mathbf{G}((p_1\ \&\ p_2) -> \mathbf{X}(\mathbf{F}(p_3\ \&\ p_4)))$\\
$\&\ \mathbf{G}(p_3 -> p_4)$
}
\\

\midrule

Two Robots Stack Cube &
\makecell[l]{
$p_1$: blue cube reaches target position\\
$p_2$: red cube grasped\\
$p_3$: red cube placed on blue cube\\
$p_4$: blue and red cubes aligned\\
$p_5$: red cube above blue cube
}
&
\makecell[l]{
$\mathbf{G}(\mathord{!}p_3 -> (\mathord{!}p_3\ \mathbf{U}\ p_1))$\\
$\&\ \mathbf{G}(p_1 -> \mathbf{X}(\mathbf{F}p_2))$\\
$\&\ \mathbf{G}(p_2 -> \mathbf{X}(\mathbf{F}(p_3\ \&\ p_4\ \&\ p_5)))$\\
$\&\ \mathbf{G}(p_3 -> (p_1\ \&\ p_4\ \&\ p_5))$
}
\\

\midrule

Camera Alignment &
\makecell[l]{
$p_1$: object reaches target position\\
$p_2$: camera reaches target height\\
$p_3$: agents grasp camera sides\\
$p_4$: camera aligned with object
}
&
\makecell[l]{
$\mathbf{G}(\mathord{!}p_4 -> (\mathord{!}p_4\ \mathbf{U}\ (p_1\ \&\ p_2\ \&\ p_3)))$\\
$\&\ \mathbf{G}((p_1\ \&\ p_2\ \&\ p_3) -> \mathbf{X}(\mathbf{F}p_4))$\\
$\&\ \mathbf{G}(p_4 -> (p_1\ \&\ p_2))$
}
\\

\midrule

Three Robots Stack Cube &
\makecell[l]{
$p_1$: blue cube reaches target position\\
$p_2$: red cube placed on blue cube\\
$p_3$: green cube placed on red cube\\
$p_4$: stack remains stable
}
&
\makecell[l]{
$\mathbf{G}(\mathord{!}p_3 -> (\mathord{!}p_3\ \mathbf{U}\ p_1))$\\
$\&\ \mathbf{G}(p_1 -> \mathbf{X}(\mathbf{F}p_2))$\\
$\&\ \mathbf{G}(p_2 -> \mathbf{X}(\mathbf{F}(p_3\ \&\ p_4)))$\\
$\&\ \mathbf{G}(p_3 -> (p_2\ \&\ p_4))$
}
\\

\midrule

Take Photo &
\makecell[l]{
$p_1$: object reaches target position\\
$p_2$: camera reaches target height\\
$p_3$: camera aligned with object\\
$p_4$: shutter arm near shutter\\
$p_5$: photo taken
}
&
\makecell[l]{
$\mathbf{G}(\mathord{!}p_5 -> (\mathord{!}p_5\ \mathbf{U}\ (p_1\ \&\ p_2\ \&\ p_3\ \&\ p_4)))$\\
$\&\ \mathbf{G}((p_1\ \&\ p_2\ \&\ p_3\ \&\ p_4) -> \mathbf{X}(\mathbf{F}p_5))$\\
$\&\ \mathbf{G}(p_5 -> (p_1\ \&\ p_2\ \&\ p_3\ \&\ p_4))$
}
\\

\midrule

Long Pipeline Delivery &
\makecell[l]{
$p_1$: first arm grasps shoe\\
$p_2$: shoe passed to next arm\\
$p_3$: shoe passed through pipeline\\
$p_4$: final arm places shoe\\
$p_5$: shoe reaches target location
}
&
\makecell[l]{
$\mathbf{G}(\mathord{!}p_5 -> (\mathord{!}p_5\ \mathbf{U}\ p_1))$\\
$\&\ \mathbf{G}(p_1 -> \mathbf{X}(\mathbf{F}p_2))$\\
$\&\ \mathbf{G}(p_2 -> \mathbf{X}(\mathbf{F}p_3))$\\
$\&\ \mathbf{G}(p_3 -> \mathbf{X}(\mathbf{F}(p_4\ \&\ \mathbf{F}p_5)))$\\
$\&\ \mathbf{G}(p_5 -> p_4)$
}
\\

\end{longtable}

\endgroup

\section{Task Descriptions}

\begingroup
\footnotesize
\setlength{\tabcolsep}{4pt}
\renewcommand{\arraystretch}{1.18}

\begin{longtable}{@{}p{0.13\textwidth} p{0.11\textwidth} p{0.36\textwidth} p{0.36\textwidth}@{}}
\caption{Task descriptions and target conditions.}
\label{tab:task_descriptions}\\

\toprule
\textbf{Task} & \textbf{Agent Number} & \textbf{Description} & \textbf{Target Condition} \\
\midrule
\endfirsthead

\toprule
\textbf{Task} & \textbf{Agent Number} & \textbf{Description} & \textbf{Target Condition} \\
\midrule
\endhead

\bottomrule
\endfoot

Pick Meat &
1 &
There is a piece of meat placed on the table. A robotic arm picks up the meat and lifts it to a specified height. &
The height of the meat reaches a predefined threshold.
\\

\midrule

Stack Cube &
1 &
A blue cube and a red cube are placed on the table. A robotic arm picks up the blue cube and places it on top of the red cube. &
The distance between the blue and red cubes is within a threshold, with the blue cube positioned at a greater height than the red cube.
\\

\midrule

Strike Cube &
1 &
A hammer and a cube are placed on the table. A robotic arm first identifies an optimal grasping position to pick up the hammer, then moves it to a suitable position to strike the cube. &
The hammerhead is positioned directly above the cube within a predefined distance threshold.
\\

\midrule

Lift Barrier &
2 &
A long barrier is placed on the table. Two robotic arms simultaneously grasp both ends of the barrier and lift it to a specified height. &
The barrier is elevated to the specified height while maintaining stability.
\\

\midrule

Pass Shoe &
2 &
A shoe is placed on the table. One robotic arm grasps the shoe and passes it to the other one, which then places it at the target location. &
The distance between the shoe and the target location is within a predefined threshold.
\\

\midrule

Place Food &
2 &
A pot and a kind of food are placed on the table. One robotic arm lifts the pot's lid, while the other picks up the food and places it inside the pot. &
The food is placed inside the pot, with the distance between the food and the center of the pot being within a predefined threshold.
\\

\midrule

Two Robots Stack Cube &
2 &
A blue cube and a red cube are placed on the table. A robotic arm picks up the blue cube to a specified position, while the other places the red cube on top of it. &
The blue cube is within the specified threshold distance from the target position. The distance between the blue and red cubes remains within a defined threshold, with the red cube positioned at a greater height than the blue cube.
\\

\midrule

Camera Alignment &
3 &
A camera and an object are placed on the table. One robotic arm picks up the object to a specified position. The other two robotic arms grasp both sides of the camera and align it to the object. &
The camera reaches a specified height, and the object is placed at the designated position that aligns with the camera.
\\

\midrule

Three Robots Stack Cube &
3 &
A blue cube, a red cube, and a green cube are placed on the table. One robotic arm picks up the blue cube to a specified position. Another arm places the red cube on top of the blue one. The last arm places the green cube on top of the red one. &
The blue cube is positioned within the specified target range. Additionally, the red cube is successfully placed on top of the blue cube, and the green cube is positioned atop the red cube.
\\

\midrule

Take Photo &
4 &
A camera and an object are placed on the table. One robotic arm picks up the object and places it to a specified position. Another two robotic arms grasp both sides of the camera and align it to the object. The last robotic arm clicks the shutter. &
The camera reaches a specified height, and the object is placed at the designated position that aligns with the camera. Additionally, the distance between the end effector of the last robotic arm and the camera's shutter is within a certain threshold.
\\

\midrule

Long Pipeline Delivery &
4 &
A shoe is placed on the table. Three robotic arms grasp the shoe and pass it to the next robotic arm. The last robotic arm places the shoe to a specified position. &
The distance between the shoe and the target location is within a predefined threshold.
\\

\end{longtable}
\endgroup

\section{Prompt Template For LTL$_f$}

\begin{tcolorbox}[
    colback=gray!5,
    colframe=black!70,
    boxrule=0.8pt,
    arc=3pt,
    left=6pt,
    right=6pt,
    top=6pt,
    bottom=6pt,
    width=\textwidth
]

\begin{lstlisting}[
basicstyle=\ttfamily\scriptsize,
breaklines=true,
breakatwhitespace=true,
columns=fullflexible
]
task_description = """
Task: Take Photo
Agent: 4
A camera and an object are placed on the table.
One robotic arm picks up the object and places it
to a specified position. Another two robotic arms
grasp both sides of the camera and align it to the
object. The last robotic arm clicks the shutter.

Target Condition:
The camera reaches a specified height, and the
object is placed at the designated position that
aligns with the camera. Additionally, the distance
between the end effector of the last robotic arm
and the camera's shutter is within a certain threshold.
"""

system_prompt = """
You are an expert in robotics and formal methods.

Given a task description, produce ONLY valid JSON with the fields:
{
  "task": "<short name of the task>",
  "propositions": [
    "<atomic proposition 1>",
    "<atomic proposition 2>",
    "..."
  ],
  "ltl_formula": "<detailed and structured LTL formula that uses all propositions and temporal operators (U, F, G) to encode the task>"
}

The LTL must:
- Use at least one temporal operator for ordering, such as U or G.
- Show conditions before and after the main event.
- Explicitly mention all propositions in the formula.
- Use ! for negation.
- Do not include any text outside the JSON.
- No explanations, no commentary.

Return only the JSON object.
"""
\end{lstlisting}

\end{tcolorbox}

\end{document}